\documentclass[sigconf]{acmart}

\usepackage{balance}

\def\BibTeX{{\rm B\kern-.05em{\sc i\kern-.025em b}\kern-.08emT\kern-.1667em\lower.7ex\hbox{E}\kern-.125emX}}

\copyrightyear{2020} 
\acmYear{2020} 
\setcopyright{acmcopyright}
\acmConference[CIKM '20]{Proceedings of the 29th ACM International Conference on Information and Knowledge Management}{October 19--23, 2020}{Virtual Event, Ireland}
\acmBooktitle{Proceedings of the 29th ACM International Conference on Information and Knowledge Management (CIKM '20), October 19--23, 2020, Virtual Event, Ireland}
\acmPrice{15.00}
\acmDOI{10.1145/3340531.3412330}
\acmISBN{978-1-4503-6859-9/20/10}

\usepackage{multirow}
\usepackage{bbding}

\begin{document}

\fancyhead{}

\title{Speaker-Aware BERT for Multi-Turn Response Selection in Retrieval-Based Chatbots}


\author{Jia-Chen Gu$^1$, Tianda Li$^2$, Quan Liu$^{1,3}$, Zhen-Hua Ling$^1$, Zhiming Su$^3$, Si Wei$^3$, Xiaodan Zhu$^2$}
\affiliation{\institution{$^1$National Engineering Laboratory for Speech and Language Information Processing, \\ University of Science and Technology of China, Hefei, China}}
\affiliation{\institution{$^2$ECE \& Ingenuity Labs, Queen's University, Kingston, Canada}}
\affiliation{\institution{$^3$State Key Laboratory of Cognitive Intelligence, iFLYTEK Research, Hefei, China}}
\email{gujc@mail.ustc.edu.cn, tianda.li/xiaodan.zhu@queensu.ca, quanliu/zhling@ustc.edu.cn, zmsu/siwei@iflytek.com}

%

%
\begin{abstract}
  In this paper, we study the problem of employing pre-trained language models for multi-turn response selection in retrieval-based chatbots.
  A new model, named \textbf{Speaker-Aware BERT (SA-BERT)}, is proposed in order to make the model aware of the speaker change information, which is an important and intrinsic property of multi-turn dialogues.
  Furthermore, a speaker-aware disentanglement strategy is proposed to tackle the entangled dialogues.
  This strategy selects a small number of most important utterances as the filtered context according to the speakers' information in them.
  Finally, domain adaptation is performed to incorporate the in-domain knowledge into pre-trained language models.
  Experiments on five public datasets show that our proposed model outperforms the present models on all metrics by large margins and achieves new state-of-the-art performances for multi-turn response selection.
\end{abstract}

%
%

%

%

%
\maketitle

\section{Introduction}

  Chatbots aim to engage users in open-domain human-computer conversations and are currently receiving increasing attention. 
  The existing work on building chatbots includes generation-based methods and retrieval-based methods.
  The first type of methods synthesize a response with a natural language generation model \cite{DBLP:conf/aaai/SerbanSBCP16}.
  In this paper, we focus on the second type and study the problem of multi-turn response selection.
  This task aims to select the best-matched response from a set of candidates, given the context of a conversation which is composed of multiple utterances \cite{DBLP:conf/sigdial/LowePSP15,DBLP:journals/dad/LowePSCLP17,DBLP:conf/acl/WuWXZL17}.
  An example of this task is illustrated in Appendix.
  
  Previous work has kept utterances separated and performs matching within a representation-interaction-aggregation framework. 
  These methods further extended the matching and attention architectures which improved the performance on this task, such as SMN \cite{DBLP:conf/acl/WuWXZL17}, DAM \cite{DBLP:conf/acl/WuLCZDYZL18}, IMN \cite{DBLP:conf/cikm/GuLL19}, IoI\cite{DBLP:conf/acl/TaoWXHZY19} and MSN \cite{DBLP:conf/emnlp/YuanZLLZHH19}. 
  We elaborate these related work in Appendix. 
  Recently, pre-trained language models have shown to achieve state-of-the-art performance on a wide range of NLP tasks \cite{DBLP:conf/naacl/DevlinCLT19}.
  \cite{DBLP:conf/acl/HendersonVGCBCS19} made the first attempt to employ pre-trained models for multi-turn response selection.
  It adopted a simple strategy by concatenating the context utterances and the response literally, and then sending them into the model for classification.
  However, this shallow concatenation has three main drawbacks.
  First, it neglects the fact that speakers are always changing in turn as a conversation progresses.
  Second, it weakens the relationships between the context utterances as they are organized in the chronological order.
  Third, due to the maximum sequence length limit (e.g., 512 for BERT-Base), pre-trained language models are unable to tackle sequences that are composed of thousands of tokens, which, however, is a typical setup in multi-turn conversations.

  In this paper, we attempt to employ the pre-trained language model and adjust it to fit the task of multi-turn response selection, in which BERT \cite{DBLP:conf/naacl/DevlinCLT19} is adopted as the basis of our work.
  We propose a new model, named \textbf{Speaker-Aware BERT (SA-BERT)}.
  First, to make the pre-trained language model aware of the speaker change information during the conversation, the model is enhanced by adding the \emph{speaker embeddings} to the token representation and adding the special segmentation tokens between the context utterances.
  These two strategies are designed to improve the conversation understanding capability of multi-turn dialogue systems.
  Furthermore, to tackle the entangled dialogues which are mixed with multiple conversation topics and are composed of hundreds of utterances, we propose a heuristic speaker-aware disentanglement strategy, which helps to select a small number of most important utterances according to the speaker information in them.
  Finally, domain adaptation is designed to incorporate specific in-domain knowledge into pre-trained language models.
  We perform the adaptation process with the same domain but different sets under the same setting.
  We can conclude that adaptation on a domain-specific corpus can help to incorporate more domain-specific knowledge, and the more similar to the task this adaptation corpus is, the more improvement it can help to achieve.

  We test our model on five datasets, Ubuntu Dialogue Corpus V1 \cite{DBLP:conf/sigdial/LowePSP15}, Ubuntu Dialogue Corpus V2 \cite{DBLP:journals/dad/LowePSCLP17}, Douban Conversation Corpus \cite{DBLP:conf/acl/WuWXZL17}, E-commerce Dialogue Corpus \cite{DBLP:conf/coling/ZhangLZZL18}, and DSTC 8-Track 2-Subtask 2 Corpus \cite{DBLP:journals/corr/abs-1911-06394}.
  Experimental results show that the proposed model outperforms the existing models on all metrics by large margins.
  Specifically,
    5.5\% $\textbf{R}_{10}@1$ on Ubuntu Dialogue Corpus V1,
    5.9\% $\textbf{R}_{10}@1$ on Ubuntu Dialogue Corpus V2,
    3.2\% \textbf{MAP} and 2.7\% \textbf{MRR} on Douban Conversation Corpus,
    8.3\% $\textbf{R}_{10}@1$ on E-commerce Corpus,
    and 15.5\% $\textbf{R}_{100}@1$ on DSTC 8-Track 2-Subtask 2 Corpus,
  leading to new state-of-the-art performances for multi-turn response selection.

  In summary, our contributions in this paper are three-fold:
  \begin{itemize}
    \item [1)]
    A new model, named Speaker-Aware BERT (SA-BERT), is designed by employing speaker embeddings and speaker-aware disentanglement strategy, to make BERT aware of the speaker change information as the conversation progresses.
    \item [2)]
    We make further analysis on the effect of adaptation to the performance of response selection.
    \item [3)]
    Experimental results show that our model achieves new state-of-the-art performances on five datasets for multi-turn response selection.
  \end{itemize}

\section{Methodology}

  Given a dialogue dataset $\mathcal{D}$, an example of the dataset is denoted as $(c,r,y)$, where $c = \{u_1,u_2,...,u_n\}$ represents a conversation context with $\{u_k\}_{k=1}^n$ as the utterances,
  $r$ is a response candidate, and $y \in \{0,1\}$ denotes a label.
  Specifically, $y=1$ indicates that $r$ is a proper response for $c$; otherwise $y=0$.
  Our goal is to learn a matching model $g(c,r)$ by minimizing a cross-entropy loss function from $\mathcal{D}$.
  For any context-response pair $(c,r)$, $g(c,r)$ measures the matching degree between $c$ and $r$.

  We present here our proposed model, named \textbf{Speaker-Aware BERT (SA-BERT)}, and a visual architecture of our input representation is illustrated in Appendix. 
  We omit an exhaustive background description of BERT. 
  Readers can refer to \cite{DBLP:conf/naacl/DevlinCLT19} for details. 

  \subsection{Speaker Embeddings \& Segmentations}


    In order to distinguish utterances in a context and model the speaker change in turn as the conversation progresses, we use two strategies to construct the input sequence for multi-turn response selection as follows.

    First, in order to model the speaker change, we propose to add additional \emph{speaker embeddings} to token representations.
    The embedding functions as indicating the speaker's identity for each utterance. 
    For conversations with two speakers, two speaker embedding vectors need to be estimated during the training process.
    The first vector is added to each token of utterances of the first speaker. 
    When the speaker changes, the second vector is employed.
    This is performed alternatively and can be extended to conversations with more speakers.

    Second, empirical results in \cite{DBLP:journals/corr/abs-1802-02614} show that segmentation tokens play an important role for multi-turn response selection. 
    To model conversation, it is natural to extend that to further model turns and utterances. 
    In this work we propose and  empirically show that using an \texttt{[EOU]} token at the end of an utterance and an \texttt{[EOT]} token at the end of a turn model interactions between utterances in a context implicitly and improve the performance consistently. 
  
  \subsection{Speaker-Aware Disentanglement Strategy}
    When more than two speakers are communicating in a common channel, there are often multiple conversation topics occurring concurrently.
    In terms of a specific conversation topic, utterances relevant to it are useful and other utterances could be considered as noise for them.
    Note that BERT is not good at dealing with sequences which are composed of more tokens than the limit (i.e., length of time steps is set to be 512).
    In order to select a small number of most important utterances, in this paper, we propose a heuristic speaker-aware disentanglement strategy as follows.

    First, we define the speaker who is uttering an utterance as the \emph{spoken-from} speaker, and define the speaker who is receiving an utterance as the \emph{spoken-to} speaker.
    Each utterance usually has the labels of both \emph{spoken-from} and \emph{spoken-to} speakers, which can be extracted from the utterance itself. 
    But some utterances may have only the \emph{spoken-from} speaker label while the \emph{spoken-to} speaker is unknown which is set to \emph{None} in our experiments.
    Second, given the \emph{spoken-from} speaker of the response, we select the utterances which have the same \emph{spoken-from} or \emph{spoken-to} speaker as the \emph{spoken-from} speaker of the response.
    Third, these selected utterances are then organized in their original chronological order and used to form the filtered context.
    Finally, the utterances selected according to their \emph{spoken-from} or \emph{spoken-to} speaker labels are assigned with the two speaker embedding vectors respectively.
    
    \begin{table}[b]
      \caption{Statistics of the datasets that our model is tested on.}
      \centering
      \begin{tabular}{c|c|ccc}
      \toprule
      \multicolumn{2}{c|}{Dataset}                      & Train & Valid & Test \\
      \hline
      \multirow{2}*{Ubuntu V1}    & pairs               & 1M    & 0.5M  & 0.5M \\
                                  & positive:negative   & 1: 1  & 1: 9  & 1: 9 \\
      \hline
      \multirow{2}*{Ubuntu V2}    & pairs               & 1M    & 195k  & 189k \\
                                  & positive:negative   & 1: 1  & 1: 9  & 1: 9 \\
      \hline
      \multirow{2}*{Douban}       & pairs               & 1M    & 50k   & 10k  \\
                                  & positive:negative   & 1: 1  & 1: 1  & 1.2: 8.8 \\
      \hline
      \multirow{2}*{E-commerce}   & pairs               & 1M    & 10k   & 10k  \\
                                  & positive:negative   & 1: 1  & 1: 1  & 1: 9 \\
      \hline
      \multirow{2}*{DSTC 8}       & pairs               & 11M   & 1M    & 1M    \\
                                  & positive:negative   & 1: 99 & 1: 99 & 1: 99 \\
      \bottomrule
      \end{tabular}
      \label{tab1}
    \end{table}
    
    \begin{table*}[!hbt]
      \caption{Evaluation results of SA-BERT and previous methods on the Ubuntu Dialogue Corpus V1 and V2.}
      \centering
      \begin{tabular}{l|c|c|c|c|c|c|c|c}
      \toprule
                             & \multicolumn{4}{c|}{Ubuntu Corpus V1} & \multicolumn{4}{c}{Ubuntu Corpus V2} \\
      \hline
                             & $\textbf{R}_2@1$ & $\textbf{R}_{10}@1 $ & $\textbf{R}_{10}@2 $ & $\textbf{R}_{10}@5 $ & $\textbf{R}_2@1$ & $\textbf{R}_{10}@1 $ & $\textbf{R}_{10}@2 $ & $\textbf{R}_{10}@5 $\\
%
      \hline
       SMN  \cite{DBLP:conf/acl/WuWXZL17}                 & 0.926 & 0.726 & 0.847 & 0.961 & -     & -     & -     & -     \\
       DUA  \cite{DBLP:conf/coling/ZhangLZZL18}           & -     & 0.752 & 0.868 & 0.962 & -     & -     & -     & -     \\
       DAM  \cite{DBLP:conf/acl/WuLCZDYZL18}              & 0.938 & 0.767 & 0.874 & 0.969 & -     & -     & -     & -     \\
       MRFN \cite{DBLP:conf/wsdm/TaoWXHZY19}              & 0.945 & 0.786 & 0.886 & 0.976 & -     & -     & -     & -     \\
       IMN  \cite{DBLP:conf/cikm/GuLL19}                  & 0.946 & 0.794 & 0.889 & 0.974 & 0.945 & 0.771 & 0.886 & 0.979 \\
       IoI  \cite{DBLP:conf/acl/TaoWXHZY19}               & 0.947 & 0.796 & 0.894 & 0.974 & -     & -     & -     & -     \\
       MSN  \cite{DBLP:conf/emnlp/YuanZLLZHH19}           & -     & 0.800 & 0.899 & 0.978 & -     & -     & -     & -     \\
      \hline
       BERT                                               & 0.950 & 0.808 & 0.897 & 0.975 & 0.950 & 0.781 & 0.890 & 0.980 \\
       \textbf{SA-BERT}                                   & \textbf{0.965} & \textbf{0.855} & \textbf{0.928} & \textbf{0.983} & \textbf{0.963} & \textbf{0.830} & \textbf{0.919} & \textbf{0.985}  \\   			
      \bottomrule
      \end{tabular}
      \label{tab2}
    \end{table*}

    \begin{table*}[!hbt]
      \caption{Evaluation results of SA-BERT and previous methods on the Douban Corpus and E-commerce Corpus.}
      \centering
      \begin{tabular}{l|c|c|c|c|c|c|c|c|c}
      \toprule
                             & \multicolumn{6}{c|}{Douban Conversation Corpus} & \multicolumn{3}{c}{E-commerce Corpus} \\
      \hline
                             & \textbf{MAP} & \textbf{MRR} & $\textbf{P}@1$ & $\textbf{R}_{10}@1 $ & $\textbf{R}_{10}@2 $ & $\textbf{R}_{10}@5 $ & $\textbf{R}_{10}@1 $ & $\textbf{R}_{10}@2 $ & $\textbf{R}_{10}@5 $ \\
      \hline
       SMN  \cite{DBLP:conf/acl/WuWXZL17}         & 0.529 & 0.569 & 0.397 & 0.233 & 0.396 & 0.724 & 0.453 & 0.654 & 0.886  \\
       DUA  \cite{DBLP:conf/coling/ZhangLZZL18}   & 0.551 & 0.599 & 0.421 & 0.243 & 0.421 & 0.780 & 0.501 & 0.700 & 0.921  \\
       DAM  \cite{DBLP:conf/acl/WuLCZDYZL18}      & 0.550 & 0.601 & 0.427 & 0.254 & 0.410 & 0.757 & -     & -     & -      \\
       MRFN \cite{DBLP:conf/wsdm/TaoWXHZY19}      & 0.571 & 0.617 & 0.448 & 0.276 & 0.435 & 0.783 & -     & -     & -      \\
       IMN  \cite{DBLP:conf/cikm/GuLL19}          & 0.570 & 0.615 & 0.433 & 0.262 & 0.452 & 0.789 & 0.621 & 0.797 & 0.964  \\
       IoI  \cite{DBLP:conf/acl/TaoWXHZY19}       & 0.573 & 0.621 & 0.444 & 0.269 & 0.451 & 0.786 & 0.563 & 0.768 & 0.950  \\
       MSN  \cite{DBLP:conf/emnlp/YuanZLLZHH19}   & 0.587 & 0.632 & 0.470 & 0.295 & 0.452 & 0.788 & 0.606 & 0.770 & 0.937  \\
      \hline
       BERT                                       & 0.591 & 0.633 & 0.454 & 0.280 & 0.470 & 0.828 & 0.610 & 0.814 & 0.973  \\
       \textbf{SA-BERT}                           & \textbf{0.619} & \textbf{0.659} & \textbf{0.496} & \textbf{0.313} & \textbf{0.481} & \textbf{0.847} & \textbf{0.704} & \textbf{0.879} & \textbf{0.985}  \\
      \bottomrule
      \end{tabular}
      \label{tab3}
    \end{table*}

  \subsection{Domain Adaptation}
    The original BERT is trained on a large text corpus to learn general language representations.
    To incorporate specific in-domain knowledge, adaptation on in-domain corpora are designed.
    In our experiments, we employ the training set of each dataset for domain adaptation without additional external knowledge. 
    Furthermore, domain adaptation is done by performing the multi-task learning that optimizing a combination of two loss functions: (1) a next sentence prediction (NSP) loss, and (2) a masked language model (MLM) loss \cite{DBLP:conf/naacl/DevlinCLT19}. 
    Specifically, the speaker embeddings can be pre-trained in the task of NSP.
    If there is no any adaptation processes, the speaker embeddings have to be initialized randomly at the beginning of the fine-tuning.
    Readers can refer to Appendix for more details.
  
  \subsection{Output Representation}
    The first token of each concatenated sequence is the \texttt{[CLS]} token, with its embedding being used as the aggregated representation for a context-response pair classification.
    This embedding captures the matching information between a context-response pair, which is sent into a classifier with a sigmoid output layer. 
    Finally, the classifier returns a score to denote the matching degree of this pair.

\section{Experiments}

  \subsection{Datasets}
    
    We tested SA-BERT on five public multi-turn response selection datasets, Ubuntu Dialogue Corpus V1 \cite{DBLP:conf/sigdial/LowePSP15}, Ubuntu Dialogue Corpus V2 \cite{DBLP:journals/dad/LowePSCLP17}, Douban Conversation Corpus \cite{DBLP:conf/acl/WuWXZL17}, E-commerce Dialogue Corpus \cite{DBLP:conf/coling/ZhangLZZL18} and DSTC 8-Track 2-Subtask 2 Corpus \cite{DBLP:journals/corr/abs-1911-06394}.
    The first four datasets have been disentangled in advance by their publishers and our proposed speaker-aware disentanglement strategy is applied to only the last DSTC 8-Track 2-Subtask 2 Corpus. 
    Some statistics of these datasets are provided in Table~\ref{tab1}.
    Readers can refer to Appendix for more details of datasets.

  \subsection{Evaluation Metrics}
    We used the same evaluation metrics as those used in previous work \cite{DBLP:conf/sigdial/LowePSP15,DBLP:journals/dad/LowePSCLP17,DBLP:conf/acl/WuWXZL17,DBLP:conf/coling/ZhangLZZL18,DBLP:journals/corr/abs-1911-06394}.
    Each model was tasked with selecting the $k$ best-matched responses from $n$ available candidates for the given conversation context $c$, and we calculated the recall of the true positive replies among the $k$ selected responses, denoted as $\textbf{R}_n@k$. 
    In addition to $\textbf{R}_n@k$, we considered mean average precision (\textbf{MAP}), mean reciprocal rank (\textbf{MRR}) and precision-at-one ($\textbf{P}@1$), especially for the Douban corpus, following settings of previous work.

  \subsection{Experimental Results}
  
    \begin{table}
      \caption{Evaluation results of SA-BERT and ablation tests of the speaker-aware disentanglement strategy (SDS) on the DSTC 8-Track 2-Subtask 2 Corpus.}
      \centering
      \begin{tabular}{clcc}
      \toprule
       Set                 & Model                                           & \textbf{MRR} & $\textbf{R}_{100}@1 $  \\
      \midrule
      \multirow{7}*{Valid} & IMN  \cite{DBLP:conf/cikm/GuLL19}               & 0.443 & 0.322   \\
                           & IMN  \cite{DBLP:conf/cikm/GuLL19} + SDS         & 0.504 & 0.375   \\
                           & BERT                                            & 0.335 & 0.258   \\
                           & BERT + SDS                                      & 0.560 & 0.440   \\
                           & SA-BERT - SDS                                   & 0.344 & 0.265   \\
                           & \textbf{SA-BERT}                                & \textbf{0.594} & \textbf{0.477}   \\
                           & SA-BERT (Ensemble)                              & 0.611 & 0.496   \\
      \midrule
       Test                & SA-BERT (Ensemble)                              & 0.621 & 0.506   \\
      \bottomrule
      \end{tabular}
      \label{tab7}
    \end{table}

    Table~\ref{tab2}, Table~\ref{tab3} and Table~\ref{tab7} present the evaluation results of SA-BERT and previous methods on the five datasets.
    All the results except ours are from the existing literature.
    Due to previous methods did not make use of pre-trained language models, we reproduced the results of BERT baseline by fine-tuning on the training set for reference, denoted as BERT for fair comparisons.
    As we can see that, BERT has already outperformed the present models on most metrics, except $\textbf{R}_{10}@5$ on Ubuntu Dialogue Corpus V1 and $\textbf{R}_{10}@1$ on E-commerce Corpus.
    Furthermore, SA-BERT outperformed the present state-of-the-art performance by large margins of
    5.5\% $\textbf{R}_{10}@1$ on Ubuntu Dialogue Corpus V1,
    5.9\% $\textbf{R}_{10}@1$ on Ubuntu Dialogue Corpus V2,
    3.2\% \textbf{MAP} and 2.7\% \textbf{MRR} on Douban Conversation Corpus,
    8.3\% $\textbf{R}_{10}@1$ on E-commerce Corpus,
    and 15.5\% $\textbf{R}_{100}@1$ on DSTC 8-Track 2-Subtask 2 Corpus. 
    These results show the ability of SA-BERT to select the best-matched response and its compatibility across domains (system troubleshooting, social network and e-commerce), achieving a new state-of-the-art performance for multi-turn response selection.
    Readers can refer to Appendix for more training details. 
    Our code has been published to help replicate our results\footnote{https://github.com/JasonForJoy/SA-BERT}.

\section{Analysis}

  \subsection{Adaptation Corpus}
  
    \begin{table}[!hbt]
      \caption{Results on the test set of Ubuntu Corpus V2, by adapting domain with different corpora and fine-tuning all on the training set of Ubuntu Corpus V2.}
      \centering
      \begin{tabular}{lcccc}
      \toprule
       Corpus                & $\textbf{R}_2@1$ & $\textbf{R}_{10}@1 $ & $\textbf{R}_{10}@2 $ & $\textbf{R}_{10}@5 $ \\
      \midrule
       None                  & 0.950 & 0.786 & 0.890 & 0.981  \\
       DSTC8                 & 0.954 & 0.803 & 0.902 & 0.981  \\
       Ubuntu V1             & 0.961 & 0.824 & 0.914 & 0.985  \\
       \textbf{Ubuntu V2}    & \textbf{0.963} & \textbf{0.830} & \textbf{0.919} & \textbf{0.985}  \\
      \bottomrule
      \end{tabular}
      \label{tab4}
    \end{table}
    
    We make some further analysis on the effect of adaptation corpus to the performance of multi-turn response selection.
    We performed the adaptation process with the same domain but different sets.
    Here, three different sets of Ubuntu were employed: DSTC 8-Track 2, Ubuntu Dialogue Corpus V1, and Ubuntu Dialogue Corpus V2.
    And then the fine-tuning process was all performed on the training set of Ubuntu Dialogue Corpus V2.
    The results on the test set of Ubuntu Dialogue Corpus V2 were shown in Table~\ref{tab4}.

    As we can see that, the adaptation process can help to improve the performance no matter which adaptation corpus was used.
    Furthermore, adaptation and fine-tuning on the same corpus achieved the best performance.
    One explanation may be that although pre-trained language models are designed to provide general linguistic knowledge, some domain-specific knowledge is also necessary for a specific task.
    Thus, adaptation on a domain-specific corpus can help to incorporate more domain-specific knowledge, and the more similar to the task this adaptation corpus is, the more improvement it can help to achieve.

  \subsection{Speaker Embeddings}
  
    \begin{table}[!hbt]
      \caption{Results on the test set of Ubuntu Corpus V2, by ablating the speaker embeddings (SE).}
      \centering
      \setlength{\tabcolsep}{1.5mm}{
      \begin{tabular}{cccccc}
      \toprule
       Pre-Train & SE  & $\textbf{R}_2@1$ & $\textbf{R}_{10}@1 $ & $\textbf{R}_{10}@2 $ & $\textbf{R}_{10}@5 $ \\
      \midrule
         No      & No  & 0.950 & 0.781 & 0.890 & 0.980  \\
         No      & Yes & 0.950 & 0.786 & 0.890 & 0.981  \\
         Yes     & No  & 0.961 & 0.825 & 0.915 & 0.984  \\
         \textbf{Yes} & \textbf{Yes} & \textbf{0.963} & \textbf{0.830} & \textbf{0.919} & \textbf{0.985}  \\
      \bottomrule
      \end{tabular}}
      \label{tab5}
    \end{table}

    The speaker embeddings were ablated and the results were reported in Table~\ref{tab5}.
    The first two lines discussed the situation in which the adaptation process were omitted, and the last two lines discussed the adaptation process were equipped with.
    The performance drop verified the effectiveness of speaker embeddings.

  \subsection{Speaker-Aware Disentanglement Strategy}
    To show the effectiveness of the speaker-aware disentanglement strategy, we also applied it to the existing model, such as IMN \cite{DBLP:conf/cikm/GuLL19}.
    The original IMN did not employ any disentanglement strategy and selected the last 70 utterances as the context, which achieved a performance of 32.2\% $\textbf{R}_{100}@1$.
    After employing the strategy, about 25 utterances were selected to form the context, which achieved a performance of 37.5\% $\textbf{R}_{100}@1$.
    Similar results can also be observed by employing this strategy to BERT and ablating this strategy in SA-BERT, as shown in Table~\ref{tab7}, which verified the effectiveness of the speaker-aware disentanglement strategy again.

\section{Conclusion}
  In this paper, we study the problem of employing pre-trained language models for multi-turn response selection in retrieval-based chatbots.
  A speaker-aware model and a speaker-aware disentanglement strategy are proposed.
  Experiments on five public datasets show that our proposed method achieves a new state-of-the-art performance for multi-turn response selection.
  Adjusting pre-trained language models to fit multi-turn response selection and designing new disentanglement strategies will be a part of our future work.


\bibliographystyle{ACM-Reference-Format}
\bibliography{sample-base}

\clearpage

\appendix

\section{Appendices}

\subsection{Task Definition}

  An example of this task is illustrated in Table~\ref{tab6}.

  \begin{table}[!hbt]
    \caption{An example of the task of multi-turn response selection.}
    \centering
    \begin{tabular}{l|l}
    \toprule
    \multicolumn{2}{l}{\textbf{Conversation}} \\
    \midrule
    \textbf{Human}  & How are you doing?  \\
    \textbf{Chatbot}& I am going to hold a drum class in Shanghai.  \\
                    & Anyone wants to join? \\
    \textbf{Human}  & Interesting! Do you have coaches who can \\
                    & help me practice drum? \\
    \textbf{Chatbot}& Of course.    \\
    \textbf{Human}  & Can I have a free first lesson? \\
    \midrule
    \multicolumn{2}{l}{\textbf{Response Candidates}} \\
    \midrule
    \textbf{Chatbot}& Sure. Have you ever played drum before? \CheckmarkBold \\
    \textbf{Chatbot}& What lessons do you want?  \XSolidBrush  \\
    \bottomrule
    \end{tabular}
    \label{tab6}
 \end{table}

\subsection{Related Work}

  The existing methods used to build an open domain dialogue system can be generally categorized into generation-based methods and retrieval-based methods.
  The generation-based methods synthesize a response with a natural language generation model by maximizing its generation probability given the previous conversation context.
  This approach enables the incorporation of rich context when mapping between consecutive dialogue turns \cite{DBLP:conf/aaai/SerbanSBCP16}.

  Our work belongs to the retrieval-based methods, which learn a matching model for a pair of a conversational context and a response candidate.
  This approach has the advantage of providing informative and fluent responses because they select a proper response for the current conversation from a repository by means of response selection algorithms \cite{DBLP:conf/sigdial/LowePSP15,DBLP:journals/dad/LowePSCLP17,DBLP:conf/acl/WuWXZL17,DBLP:conf/coling/ZhangLZZL18}.
  Previous work on retrieval-based chatbots focused on single-turn response selection \cite{DBLP:conf/emnlp/WangLLC13,DBLP:journals/corr/JiLL14}.
  Recently, researchers have extended the focus to the multi-turn conversation, which is more practical for real applications.
  Some earlier work on multi-turn response selection matched a response with concatenating the context utterances literally into a single long sequence, and calculating its matching score with a response candidate \cite{DBLP:conf/sigdial/LowePSP15,DBLP:journals/corr/KadlecSK15,DBLP:journals/dad/LowePSCLP17}.
  Recent work has kept utterances separated and performed matching within a representation-interaction-aggregation framework, which improved the performance on this task.
  For example, \cite{DBLP:conf/emnlp/ZhouDWZYTLY16} proposed a multi-view model, including an utterance view and a word view.
  \cite{DBLP:conf/acl/WuWXZL17} proposed the sequential matching network (SMN) which first matched the response with each utterance and then accumulated the matching information by recurrent neural network.
  \cite{DBLP:conf/coling/ZhangLZZL18} proposed the deep utterance aggregation network (DUA) which refined utterances and employed self-matching attention to route the vital information in each utterance.
  \cite{DBLP:conf/acl/WuLCZDYZL18} proposed the deep attention matching network (DAM) which constructed representations at different granularities with stacked self-attention and cross-attention.
  \cite{DBLP:conf/wsdm/TaoWXHZY19} proposed the multi-representation fusion network (MRFN) with multiple types of representations.
  \cite{DBLP:conf/cikm/GuLL19} proposed the interactive matching network (IMN) which performed the global and bidirectional interactions between the context and response.   
  \cite{DBLP:conf/acl/TaoWXHZY19} proposed the interaction over interaction (IoI) model which performed matching by stacking multiple interaction blocks.
  \cite{DBLP:conf/emnlp/YuanZLLZHH19} proposed the multi-hop selector network (MSN) which utilized a multi-hop selector to select the relevant utterances as context.
  \cite{DBLP:conf/acl/HendersonVGCBCS19} made the first attempt to employ pre-trained language models for multi-turn response selection which concatenated the context utterances and the response literally and sent into the model for classification.

  \subsection{Input Representation} \label{sec2}

    \begin{figure*}
      \centering
      \includegraphics[width=17.5cm]{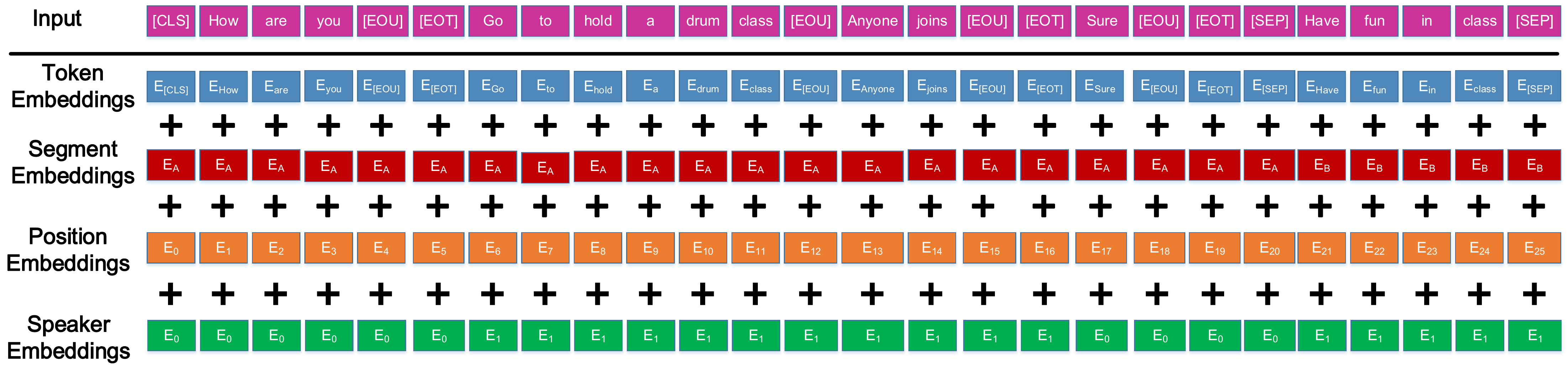}
      \caption{The input representation of SA-BERT. The final input embeddings are the sum of the token embeddings, the segmentation embeddings, the position embeddings and the speaker embeddings.}
      \label{fig1}
    \end{figure*}

    A visual architecture of our input representation is illustrated in Figure~\ref{fig1}.

  \subsection{Adaptation Tasks}  \label{sec3}
    Here, the masked language model (MLM) and the next sentence prediction (NSP) \cite{DBLP:conf/naacl/DevlinCLT19} are employed.

    \paragraph{MLM}
    We follow the experimental settings in the original BERT by masking some percentage of the input tokens at random and then predicting only those masked tokens to train a deep bidirectional representation.
    In more detail, we replace the word with the \texttt{[MASK]} token at 80\% of the time, with a random word at 10\% of the time, and with the original word at 10\% of the time.

    \paragraph{NSP}
    If there is no pre-training process, the speaker embeddings have to be initialized at random at the beginning of the fine-tuning process.
    To achieve a better performance, the speaker embeddings can be pre-trained with the help of NSP.
    Here, the sentence A and sentence B are constructed with the same method as that used in the fine-tuning process.
    The positive responses are true responses that follow the context, and the negative responses are randomly sampled.
    The embedding of the \texttt{[CLS]} token is used as the aggregated representation for classification.

  \subsection{Datasets}  \label{sec4}

    We tested SA-BERT on five public multi-turn response selection datasets, Ubuntu Dialogue Corpus V1 \cite{DBLP:conf/sigdial/LowePSP15}, Ubuntu Dialogue Corpus V2 \cite{DBLP:journals/dad/LowePSCLP17}, Douban Conversation Corpus \cite{DBLP:conf/acl/WuWXZL17}, E-commerce Dialogue Corpus \cite{DBLP:conf/coling/ZhangLZZL18} and DSTC 8-Track 2-Subtask 2 Corpus \cite{DBLP:journals/corr/abs-1911-06394}.
    The first four datasets have been disentangled in advance by their publishers and our proposed speaker-aware disentanglement strategy is applied to only the last DSTC 8-Track 2-Subtask 2 Corpus. 
    Here, we adopted the version of Ubuntu Dialogue Corpus V1 shared in \citet{DBLP:journals/corr/XuLWSW16}, in which numbers, paths and URLs were replaced by placeholders. 
    Compared with Ubuntu Dialogue Corpus V1, the training, validation and test dialogues in the V2 dataset were generated in different periods without overlap. 
    In the DSTC 8-Track 2-Subtask 2 Corpus, the candidate pool may not contain the correct response, so we need to choose a threshold.
    When the probability of positive labels was smaller than the threshold, we predicted that candidate pool did not contain the correct response.
    The threshold was selected among [0.6, 0.65, .., 0.95] based on the validation set.
    In all of the Ubuntu corpora, the positive responses are true responses from humans, and the negative responses are randomly sampled.
    The Douban Conversation Corpus was crawled from a Chinese social network on open-domain topics.
    It was constructed in a similar way to the Ubuntu corpus.
    The Douban Conversation Corpus collected responses via a small inverted-index system, and labels were manually annotated.
    The Douban Conversation Corpus is different from the other three datasets in that it includes multiple correct candidates for a context in the test set, which leads to low $\textbf{R}_{n}@k$, e.g., if there are 3 correct responses, the maximum $\textbf{R}_{10}@1$ is 0.33.
    Hence, $\textbf{MAP}$ and $\textbf{MRR}$ are recommended for reference.
    The E-commerce Dialogue Corpus collected real-world conversations between customers and customer service staff from the largest e-commerce platform in China.
    The DSTC 8-Track 2-Subtask 2 Corpus does not release the labels of the test set.
    Participants should submit their results on the test set to the official and then be evaluated by them.
    Thus, we submitted only one result to the official and we provide other results on the validation set for reference.

  \subsection{Training Details}
    Most hyper-parameters of the original BERT were followed \cite{DBLP:conf/naacl/DevlinCLT19} except the following configurations.
    The initial learning rate was set to 2e-5 and was linearly decayed by L2 weight decay.
    The maximum sequence length of the concatenation of a context-response pair was set to 512.
    The training batch size was set to 25.
    The maximum number of training epochs was set to 3.
    We used the validation set to set the stop condition in order to select the best model for testing.
    All codes were implemented in the TensorFlow framework \cite{DBLP:conf/osdi/AbadiBCCDDDGIIK16} and have be published to help replicate our results.


\end{document}